\documentclass[letterpaper]{article}
\usepackage{aaai16}
\usepackage{times}
\usepackage{helvet}
\usepackage{courier}
\usepackage{graphics}
\usepackage{graphicx}
\usepackage{subfigure}

\usepackage{mathrsfs}
\usepackage{amsfonts}
\usepackage{amsmath}
\usepackage{enumitem}
\usepackage{amsthm, bm, amssymb}

\begin{document}
\title{Learning Lexical Entries for Robotic Commands using Crowdsourcing}
\author{Junjie Hu, Jean Oh, Anatole Gershman\\
School of Computer Science, Carnegie Mellon University, Pittsburgh, PA 15213\\
junjieh@cs.cmu.edu, jeanoh@nrec.ri.cmu.edu, anatoleg@cs.cmu.edu\\
}
\maketitle 

% --------------------------------------------
\begin{abstract}
Robotic commands in natural language usually contain various spatial descriptions that are semantically similar but syntactically different.  Mapping such syntactic variants into semantic concepts that can be understood by robots is challenging due to the high flexibility of natural language expressions. To tackle this problem, we collect robotic commands for navigation and manipulation tasks using crowdsourcing.  We further define a robot language and use a generative machine translation model to translate robotic commands from natural language to robot language.  The main purpose of this paper is to simulate the interaction process between human and robots using crowdsourcing platforms, and investigate the possibility of translating natural language to robot language with paraphrases.  %By jointly learning the spatial relationship across two different but similar task domains, we demonstrate the effectiveness of our method in improving the grounding performance.
%Learning a semantic lexicon is often a fundamental first step in building a human machine interaction system that can interpret the meaning of natural language. It is especially important in machine teaching where a human partner teaches an unseen semantic lexicon to an intelligent machine.  To interactively learn an unseen semantic lexicon, we propose an active learning approach for semantic lexicon learning, which allows a machine to perform semantic reasoning over a sentence and actively select ambiguous lexicon to enquiry its human partner.  By exploiting the extracted semantic grammars during the human-machine interaction, our designed system can gradually mimic its human partner and expressively generate natural language description with rich semantic meaning. 
\end{abstract}
% --------------------------------------------

% --------------------------------------------
\section{Introduction}
\label{sec:introduction}
%\textbf{Importance of phrase-based lexicons in Robotic Commands
%\textbf{(Lexicon Learning and Semantic Grammar Extraction)} 
Natural language provides an efficient way for untrained human to instruct a robot to perform collaborative tasks, e.g., navigation and manipulation. However, learning to interpret the meaning of natural language commands is a challenging task~\cite{dukes2014semeval,DBLP:conf/aaai/PereraA13,DBLP:conf/aaai/ChenM11}, especially when the robot has little or no prior knowledge of the phrasal expressions in natural language.  Due to high flexibility of natural language, it is non-trivial for a robot to cover all the phrasal expressions in natural language when its interpretation module is initially built. % Furthermore, robotic commands usually specify spatial information on objects in the environment which is crucial to perform semantic grounding on objects. 

%Training a language model to interpret the meaning of natural language commands requires lots of annotated data.  
Popular crowdsourcing platforms such as Amazon Mechanical Turk, provide a fast and cheap way to collect interactive data from participants in a wide range of different communities.  Hence, simulating the human machine interaction process for information extraction on crowdsourcing platforms has attracted lots of research interests~\cite{DBLP:conf/hcomp/NguyenWL15,DBLP:conf/hcomp/HladkaHL14,DBLP:conf/hcomp/GoldbergWK13}.  To encourage the diversity of robotic commands, we simulate the interactive process between a robot and various untrained users on Amazon Mechanical Turk, and collect robotic commands during the process. %More specifically, we show an image that depicts a robotic task, and ask the turker to give a corresponding commands for instructing the robot to finish the task.  
We further apply a phrase-based machine translation model to mapping natural language command to a robotic language that can be understood by a robot. 

% --------------------------------------------

% --------------------------------------------
%\input{related-work.tex}
% --------------------------------------------

% --------------------------------------------
\section{Phrase-based Machine Translation Model}\label{sec:system-overview}
To tackle the problem of translating natural language commands to language that can be understood by robots, we first define a robot language that consists of predefined key concepts in the robotic task domains.  For example, in the navigation task domain, we define the following key concepts.  
\begin{itemize}[topsep=0ex,itemsep=-1ex,partopsep=0ex,parsep=1ex]
\item Action:= navigate
\item Object:= traffic barrel $|$ building $|$ car
\item Relation:= left $|$ right $|$ front $|$ back
\end{itemize}
Each robot language command can be deterministically constructed by a combination of key concepts in the task domains. See Figure~\ref{fig:navigation-example} for an illustration.  We then adapt a phrase-based machine translation model to translate robotic commands from natural language to the robot language.  For the phrase-based machine translation model, the key component is the extracted phrase table that stores several lexical entries.  For a particular input (source-language) sentence $s=s_1\cdots s_n$, each lexical entry is defined as a tuple $(b,e,r)$, specifying that the span $s_b\cdots s_e$ in the source-language sentence can be translated as the target-language string $r$. For each lexical entry $p=(b,e,r)$, we estimate a score $g(p)\in\mathbb{R}$ that measures the likelihood of translating the span to the target language string by relative frequency under the translation model.  For a given lexicon entry $p$, $b(p), e(p), r(p)$ denote its three components respectively.  A derivation $y$ of a source-language sentence is defined as a finite sequence of phrases, $p_1, p_2\cdots p_L$.  For any derivation $y$, $r(y)$ refers to the translation sentence constructed by concatenating the strings $r(p_1), r(p_2),\cdots r(p_L)$. For a source-language sentence $s$, we denote $\mathcal{Y}(s)$ as a set of possible derivations of $s$.  %we denote $\mathcal{Y}(s)$ as a set of possible derivations of $s$.we denote $\mathcal{Y}(s)$ as a set of possible derivations of $s$.we denote $\mathcal{Y}(s)$ as a set of possible derivations of $s$.we denote $\mathcal{Y}(s)$ as a set of possible derivations of $s$.

Based on the above notations, we aim to extract lexical entries from parallel textual corpus collected on crowdsourcing platforms, and seek the optimal derivation $y^*$ using beam search for the maximum derivation score $f(y^*)$ among all possible derivations $\mathcal{Y}(s)$ under a phrase-based translation model.  

In Equation~\ref{eq:f(y)}, the score $f(y)$ of a derivation $y$ consists of three parts: (1) $h(r(y))$ is the log-probability of the target string $r(y)$ under a smoothed trigram language model; (2) $g(p_k)$ is the score of $p_k$ under a translation model; (3) $|e(p_k)+1-b(p_{k+1})|$ is the distortion penalty for reordering word alignments between source and target languages. 
\small 
\begin{align}\label{eq:f(y)}
f(y) = w_h h(r(y)) + w_g\sum_{k=1}^{L}g(p_k) + w_d\sum_{k=1}^{L-1}|e(p_k)+1-b(p_{k+1})|
\end{align}
\normalsize
where $w_h$, $w_g$ and $w_d$ are the weights of the scores given by the language model, the translation model and the distortion penalty respectively.  Hence the optimal derivation of a source-language sentence $s$ can be obtained by $\arg\max_{y\in\mathcal{Y}(s)} f(y)$.
\section{Experiment}
\label{sec:experiment}
We present the process of collecting experimental data on Amazon Mechanical Turk, a popular crowdsourcing platform, and extract parallel lexical entries using Moses~\cite{koehn2007moses}, a machine translation tool.
\subsection{Stimulation and Data Collection}
%By showing an image that depicts the behaviour of a robot, the turker is asked to give two natural language commands (denoted as $s$ and $t$) respectively before and after showing the turker the robotic concepts in several drop-down lists that can be understood by the robot.  The turker is asked to select the correct robotic concepts which can be used to construct a robotic command (denoted as $r$) for the same image.  Totally we collect 80 tuples of $(s,t,r)$ for navigation task and 120 tuples of $(s,t,r)$ tuples for manipulation task.
By showing an image that depicts the behaviour of a robot, a turker is first asked to give a command in English (denoted as $s$) that clearly indicates the spatial information between objects in the environment for a robot.  Next, the turker is shown some robotic concepts in several drop-down lists, and asked to select the correct robotic concepts that can be used to construct a robotic command (denoted as $r$) for the same image. Finally we simulate the scenario where the robot can actively ask for a paraphrase sentence (denoted as $t$) of the robotic command $r$ in order to help it understand $s$.  Totally we collect 88 tuples of $(s,t,r)$ for navigation task and 120 tuples of $(s,t,r)$ tuples for manipulation task.

\subsection{Phrasal Lexicon Extraction and Translation}
To investigate the possibility of using paraphrase sentences to enhance the phrase-based machine translation, we first use Moses to extract parallel phrases between $s$ and $r$.  Then we use Moses to extract parallel phrases between $t$ and $r$.  Table~\ref{tab:cnt} shows the total number of extracted lexical entries when we translate from $s$ to $r$ and from $t$ to $r$.  Comparing the second column with the third one in Table~\ref{tab:cnt}, we observe that more lexical entries are extracted from parallel sentences between $t$ and $r$ than those between $s$ and $r$.  This convinces our idea that turkers usually paraphrase natural language commands that are more semantically closed to the robot language commands after the robotic concepts are shown to them.  Table~\ref{tab:phrase} shows some lexical entries extracted from natural language commands $t$ paired with robot language commands $r$.  We observe that the extracted lexical entries capture the similarity between source-language phrases and target-language phrases, thus enabling many-to-one mapping from syntactic variants in natural language to unique robotic concepts.  %Moreover, it is also possible to use alternative estimation of the phrasal score $g(p_k)$ in Equation~\re on enhancing the translation from $s$ to $r$ via using lexicals extracted from $t$ to $r$.

\begin{table}[thb]
\vskip -0.2in
\centering
\caption{Number of extracted lexical entries}
\label{tab:cnt}
%\begin{center}
\begin{tabular}{|c|c|c|}
\hline
             & \#phrase from ($s,r$) & \#phrase from ($t,r$) \\ \hline
Navigation   & 160               & 748               \\ \hline
Manipulation & 128               & 298               \\ \hline
\end{tabular}
%\end{center}
\vskip -0.1in
\end{table}

\begin{table}[thb]
\vskip -0.2in
\centering
\caption{Examples of extracted lexical entries}
\label{tab:phrase}
%\begin{center}
\begin{tabular}{|c|c|}
	\hline
	\multicolumn{2}{|c|}{Navigation Task}\\
	\hline
	Natural Language & Robot Language \\
	\hline
	go straight until you reach a car & navigate to the car\\
	backyard of the building & behind the building\\
	find the car & to the car \\
	which stands before & that is in front\\
	move forward to & navigate to\\
	located at the right hand side of &is on the right of\\
	\hline
\end{tabular}
%\end{center}
\end{table}

\begin{figure}[ht]
\vskip -0.1in
\begin{center}
\centerline{
\subfigure[$$]{\includegraphics[width=.181\textwidth]{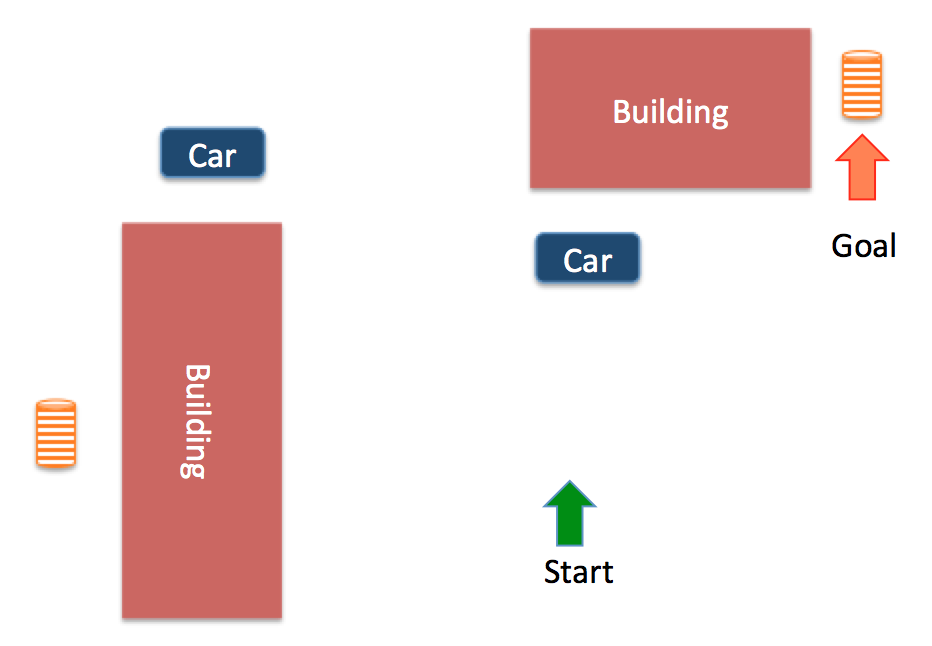}\label{fig:synthetics_global}}
\subfigure[$$]{\includegraphics[width=.181\textwidth]{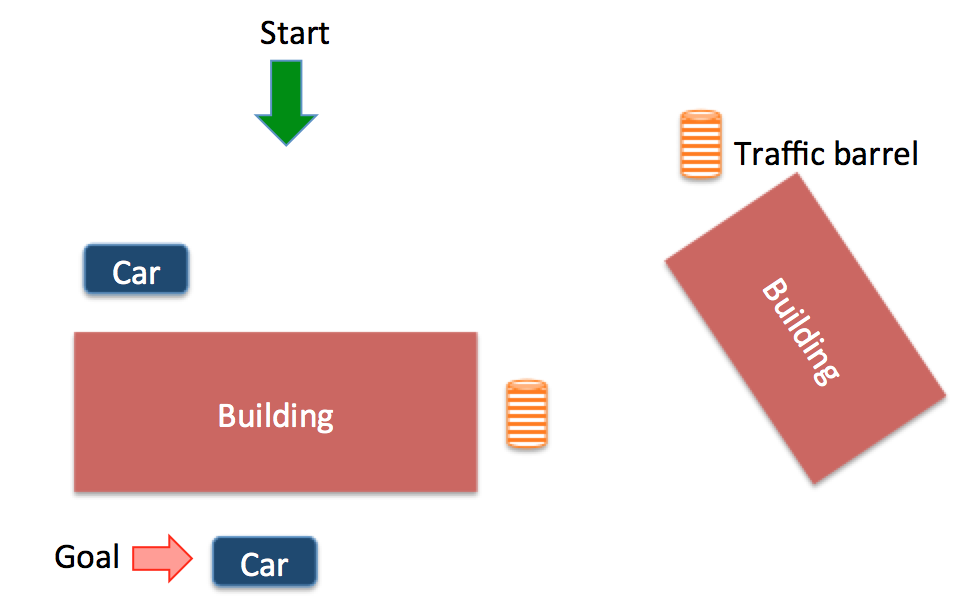}\label{fig:synthetics_global}}
}
\vskip -0.2in
\caption{Navigation examples: (a) \textbf{navigate (Action)} to the \textbf{traffic barrel (Object)} that is on the \textbf{right (Relation)} of the \textbf{building (Object)}; (b) \textbf{navigate (Action)} to the \textbf{car (Object)} that is on the \textbf{back (Relation)} of the \textbf{building (Object)}}
\label{fig:navigation-example}
\end{center}
\vskip -0.4in
\end{figure}

%\begin{figure}
%\centering
%\includegraphics[width=0.6\linewidth]{./figure/navigation-right}
%\caption{robot language command: \textbf{navigate (Action)} to the \textbf{traffic barrel (Target Object)} that is on the \textbf{right (Relation)} of the \textbf{building (Reference Object)}.}
%\label{fig:navigation-example}
%\end{figure}

\begin{table}[tb]
\vskip -0.3in
\centering
\caption{Examples of phrase-based translation}
\label{tab:translation}
%\begin{center}
\begin{tabular}{|c|c|}
	\hline
	\multicolumn{2}{|c|}{Navigation Task}\\
	\hline
	Natural Language & Translated Robot Language \\
	\hline
	\begin{tabular}[c]{@{}l@{}}go to the traffic barrel that \\ is located on the right hand\\ side of the building\end{tabular} &\begin{tabular}[c]{@{}l@{}}\textbf{navigate (Action)} to the \\ \textbf{traffic barrel (Object)} that\\is on the \textbf{right (Relation)}\\of the \textbf{building (Object)}\end{tabular} \\
	\hline
	\begin{tabular}[c]{@{}l@{}}go straight forward until you\\reach the building. go to the\\ car behind the building.\end{tabular} &\begin{tabular}[c]{@{}l@{}}\textbf{navigate (Action)} to the \\ \textbf{building (Object)} that is \\ \textbf{navigate (Action)} to the \\ \textbf{car (Object)} that is \textbf{behind} \\ \textbf{(Relation)} the \textbf{building} \\\textbf{(Object)}\end{tabular} \\
	\hline
\end{tabular}
%\end{center}
\end{table}

By optimizing the objective function in Equation~\ref{eq:f(y)}, we generate the translated robot language sentence using the extracted lexical entries.  In Table~\ref{tab:translation}, we show two translation results of the examples used in Figure~\ref{fig:navigation-example}. In the first result, the machine translation model can successfully translate the natural language command to the correct robot language command.  While in the second result, the translation is not completely correct because the natural language command contains the detail steps for the navigation task.  Mapping detail descriptions to highly abstract robot concepts requires more sophisticated semantic reasoning over the natural language. We leave it as our future work.
%\begin{itemize}
%\item \textbf{Correct robot language}: navigate to the car the traffic barrel that is on the right of the building.
%\item \textbf{Natural language}: go to the traffic barrel that is located on the right of the building.
%\item \textbf{Translated result}: navigate to the car that is on the left of the building the car that is in front fo the building. %NAVIGATE TO THE CAR THAT IS ON THE LEFT OF THE BUILDING then till THE CAR THAT IS IN FRONT OF THE BUILDING . 
%\end{itemize}
%
%\begin{itemize}
%\item \textbf{Correct robot language}: navigate to the car the traffic barrel that is on the right of the building.
%\item \textbf{Natural language}: go to the traffic barrel that is located on the right of the building.
%\item \textbf{Translated result}: navigate to the car that is on the left of the building the car that is in front fo the building. %NAVIGATE TO THE CAR THAT IS ON THE LEFT OF THE BUILDING then till THE CAR THAT IS IN FRONT OF THE BUILDING . 
%\end{itemize}

%\begin{itemize}
%\item go to the traffic barrel that is located on the right of the building
%\item navigate to the traffic barrel that is on the right of the building.
%\end{itemize}

% --------------------------------------------

% --------------------------------------------
\section{Conclusion}\label{sec:conclusion}
In this paper, we simulate the human robot communication on Amazon Mechanical Turk and collect robotic commands for navigation and manipulation tasks using crowdsourcing.  We further investigate the possibility of bridging the gap between natural language command and robot language command using paraphrasing.  We will conduct our future work in several challenging aspects.  First, lexical entries extracted from different but similar robotic tasks can be shared across tasks.  Second, machine teaching by paraphrasing can be integrated with active learning techniques.  Robots can perform reasoning over the confusing phrases and actively ask their human partners for paraphrasing. % also interesting because it stimulates human learning process where a teacher explains the confusing phrases by paraphrasing.
% --------------------------------------------

%\section{ Acknowledgments}
\subsection*{Acknowledgments}
This work was conducted in part through collaborative participation in the Robotics Consortium sponsored by the U.S Army Research Laboratory under the Collaborative Technology Alliance Program, Cooperative Agreement W911NF-10-2-0016, and in part by ONR under MURI grant ``Reasoning in Reduced Information Spaces'' (no. N00014-09-1-1052). The views and conclusions contained in this document are those of the authors and should not be interpreted as representing the official policies, either expressed or implied, of the Army Research Laboratory of the U.S. Government. The U.S. Government is authorized to reproduce and distribute reprints for Government purposes notwithstanding any copyright notation herein.

\bibliographystyle{aaai}
\bibliography{./ref}

\begin{thebibliography}{}

\bibitem[\protect\citeauthoryear{Chen and
  Mooney}{2011}]{DBLP:conf/aaai/ChenM11}
Chen, D.~L., and Mooney, R.~J.
\newblock 2011.
\newblock Learning to interpret natural language navigation instructions from
  observations.
\newblock In {\em Proceedings of the Twenty-Fifth {AAAI} Conference on
  Artificial Intelligence, {AAAI} 2011, San Francisco, California, USA, August
  7-11, 2011}.

\bibitem[\protect\citeauthoryear{Dukes}{2014}]{dukes2014semeval}
Dukes, K.
\newblock 2014.
\newblock Semeval-2014 task 6: Supervised semantic parsing of robotic spatial
  commands.
\newblock {\em SemEval 2014} ~45.

\bibitem[\protect\citeauthoryear{Goldberg, Wang, and
  Kraska}{2013}]{DBLP:conf/hcomp/GoldbergWK13}
Goldberg, S.~L.; Wang, D.~Z.; and Kraska, T.
\newblock 2013.
\newblock {CASTLE:} crowd-assisted system for text labeling and extraction.
\newblock In {\em Proceedings of the First {AAAI} Conference on Human
  Computation and Crowdsourcing, {HCOMP} 2013, November 7-9, 2013, Palm
  Springs, CA, {USA}}.

\bibitem[\protect\citeauthoryear{Hladk{\'{a}}, Hana, and
  Luksov{\'{a}}}{2014}]{DBLP:conf/hcomp/HladkaHL14}
Hladk{\'{a}}, B.; Hana, J.; and Luksov{\'{a}}, I.
\newblock 2014.
\newblock Crowdsourcing in language classes can help natural language
  processing.
\newblock In {\em Proceedings of the Seconf {AAAI} Conference on Human
  Computation and Crowdsourcing, {HCOMP} 2014, November 2-4, 2014, Pittsburgh,
  Pennsylvania, {USA}}.

\bibitem[\protect\citeauthoryear{Koehn \bgroup et al\mbox.\egroup
  }{2007}]{koehn2007moses}
Koehn, P.; Hoang, H.; Birch, A.; Callison-Burch, C.; Federico, M.; Bertoldi,
  N.; Cowan, B.; Shen, W.; Moran, C.; Zens, R.; et~al.
\newblock 2007.
\newblock Moses: Open source toolkit for statistical machine translation.
\newblock In {\em Proceedings of the 45th annual meeting of the ACL on
  interactive poster and demonstration sessions},  177--180.
\newblock Association for Computational Linguistics.

\bibitem[\protect\citeauthoryear{Nguyen, Wallace, and
  Lease}{2015}]{DBLP:conf/hcomp/NguyenWL15}
Nguyen, A.~T.; Wallace, B.~C.; and Lease, M.
\newblock 2015.
\newblock Combining crowd and expert labels using decision theoretic active
  learning.
\newblock In {\em Proceedings of the Third {AAAI} Conference on Human
  Computation and Crowdsourcing, {HCOMP} 2015, November 8-11, 2015, San Diego,
  California.},  120--129.

\bibitem[\protect\citeauthoryear{Perera and
  Allen}{2013}]{DBLP:conf/aaai/PereraA13}
Perera, I.~E., and Allen, J.~F.
\newblock 2013.
\newblock {SALL-E:} situated agent for language learning.
\newblock In {\em Proceedings of the Twenty-Seventh {AAAI} Conference on
  Artificial Intelligence, July 14-18, 2013, Bellevue, Washington, {USA.}}

\end{thebibliography}
\end{document}